\documentclass[conference]{IEEEtran}
\IEEEoverridecommandlockouts
\usepackage{cite}

\usepackage{newtxtext,newtxmath}

\usepackage{graphicx}
\usepackage{textcomp}
\usepackage{xcolor}

\usepackage{caption}
\def\BibTeX{{\rm B\kern-.05em{\sc i\kern-.025em b}\kern-.08em
    T\kern-.1667em\lower.7ex\hbox{E}\kern-.125emX}}
\begin{document}

\title{Detecting Statistically Significant Fairness Violations in Recidivism Forecasting Algorithms*\\

}

\author{\IEEEauthorblockN{Animesh Joshi}
\IEEEauthorblockA{\textit{Purdue University} \\
San Diego, California \\
joshi197@purdue.edu}
}

\maketitle

\begin{abstract}

Machine learning algorithms are increasingly deployed in critical domains such as finance, healthcare, and criminal justice [1]. The increasing popularity of algorithmic decision-making has stimulated interest in algorithmic fairness within the academic community. Researchers have introduced various fairness definitions that quantify disparities between privileged and protected groups, use causal inference to determine the impact of race on model predictions, and that test calibration of probability predictions from the model. Existing literature does not provide a way in which to assess whether observed disparities between groups are statistically significant or merely due to chance. This paper introduces a rigorous framework for testing the statistical significance of fairness violations by leveraging k-fold cross-validation [2] to generate sampling distributions of fairness metrics. This paper introduces statistical tests that can be used to identify statistically significant violations of fairness metrics based on disparities between predicted and actual outcomes, model calibration, and causal inference techniques [1].  We demonstrate this approach by testing recidivism forecasting algorithms trained on data from the National Institute of Justice. Our findings reveal that machine learning algorithms used for recidivism forecasting exhibit statistically significant bias against Black individuals under several fairness definitions, while also exhibiting no bias or bias against White individuals under other definitions. The results from this paper underscore the importance of rigorous and robust statistical testing while evaluating algorithmic decision-making systems.

\end{abstract}

\section{Introduction}

Machine learning algorithms are being used more frequently within the criminal justice system to assist judges, probation officers, and parole boards in making decisions regarding sentencing and eligibility for parole. Recidivism risk assessment tools are machine learning algorithms that forecast the probability of a prisoner reoffending and returning to prison. These tools are often used to provide insights to judges and parole boards to help make a data-driven decision as to whether or not a prisoner should be eligible for parole or given a higher sentence. In 2016, ProPublica conducted an analysis of the COMPAS recidivism forecasting tool [11] and this analysis determined that the tool was biased against black individuals and could lead to those individuals receiving harsher sentences and losing the opportunity for parole. 
In 2021, the National Institute of Justice constructed a dataset [5] that includes observed recidivism outcomes and covariates such as demographic data, criminal history, drug abuse, prison misbehavior, and racial identity. This study uses this dataset to train statistical learning algorithms to forecast recidivism outcomes and to assess whether these models exhibit statistically significant levels of racial bias against black individuals. The study uses the fairness definitions formalized by Verma and Rubin [1] and rigorously tests these definitions to ensure that violations of these definitions are statistically significant and do not occur due to chance.
The primary contribution of this work is to provide a framework for identifying fairness violations in machine learning algorithms that ensures that the violations are statistically significant. The secondary contribution of this work is to test, using the framework, whether models trained to forecast recidivism outcomes exhibit statistically significant bias.

\subsection{Methodology}

There are four different types of machine learning metrics, metrics based on predicted and actual outcomes, metrics based on predicted probabilities and actual outcomes, metrics based on similarity of observations, and metrics based on causal graphs. Metrics based on predicted and actual outcomes examine whether there are disparities between privileged and protected groups when it comes to model statistics such as false positive rates, false positive rates, and positive prediction rates. Metrics based on predicted probabilities and actual outcomes ensure that privileged and protected individuals have equivalent probabilities to truly belong to the positive class and to ensure that among individuals with both positive and negative outcomes, individuals in the protected and privileged classes must have equivalent average probability predictions. Metrics based on similarity of observations ensure that individuals with similar covariates but different races are predicted to experience the same outcome. Metrics based on causal graphs analyze the causal graph to ensure that the variable of interest (race) does not affect the outcome.

K-Fold cross validation is a commonly used algorithm for model evaluation. In essence, a dataset is divided into K folds. Each fold is used as a test set at some point and the model is trained K times with the other K-1 folds used as training data. Model statistics are then averaged across all K test folds. This algorithm can therefore also be used to generate distributions for metrics among protected individuals and metrics among privileged individuals to enable statistical inference to detect statistically significant fairness violations.

To identify statistically significant violations of fairness metrics based on predicted and actual outcomes, distributions for model statistics (TPR, FPR, PPR, etc.) among protected and privileged individuals can be generated using K-Fold cross validation and these distributions can be used to identify statistically significant disparities. To identify statistically significant violations of fairness metrics based on predicted probabilities and actual outcomes, K-fold cross validation can be used to generate probabilistic predictions for each observation and then model calibration tests can be used to conduct statistical inference. To identify statistically significant violations of fairness metrics, K-Fold cross validation can be used to generate predictions for individuals and similar observations with different races. Randomization tests can then be used to identify whether there are statistically significant disagreements between the predictions for white individuals and their black “neighbors” and vice versa. There is no causal graph that can be drawn using current literature regarding recidivism, and thus, this paper refrains from identifying statistically significant violations of metrics based on causal graphs.

The data used for this experiment comes from the National Institute of Justice. The dataset includes ~250,000 observations with recidivism outcomes and various covariates (race, age, education levels, supervision risk scores, prior arrests, drug violations, etc.) for each observation. 250-fold cross validation is used to create 250 potential test datasets and four machine learning algorithms (Logistic Regression, Linear Discriminant Analysis, Random Forest, Extreme Gradient Boosting) are trained using each of the 250 “folds” as a test set separately. Each model is then evaluated for fairness using the approach mentioned above and the models are then compared based on fairness and overall accuracy to determine which models perform the best and which models are the most equitable.

\section{Results}

\subsection{Model Accuracy}

The table below displays the average accuracy of each of the four machine learning algorithms across the 250 unique folds. This table shows that the logistic regression model performs the worst with an accuracy score of 65.6\%, the linear discriminant analysis model achieves an accuracy score of 71.5\%, the random forest model achieves an accuracy score of 72.2\%, while the extreme gradient boosting algorithm performs the best with an accuracy score of 73.7\%.

\begin{table}[h!]

\caption{Figure 1.1}
\scalebox{1.6}{\begin{tabular}{|l|l|}
\hline
Model                        &  Accuracy \\ \hline
Logistic Regression          & 0.656    \\ \hline
Linear Discriminant Analysis & 0.715    \\\hline
Random Forest                & 0.722    \\ \hline
Gradient Boosting            & 0.737  

\\ \hline
\end{tabular}}

\end{table}

\subsection{Testing Group Fairness}

Group Fairness is a fairness metric that ensures that individuals in both the protected and privileged classes have equal rates of positive prediction (no recidivism). Mathematically, this definition tests whether $P(\hat{y}=0|race=Black) = P(\hat{y}=0|race = White)$. K-Fold cross validation helps us obtain distributions for $P(\hat{y}=0|race=Black)$ and $P(\hat{y}=0|race=White)$ where $n=250$. A two sample t-test can then be used to compare means of these distributions and test whether there is a disparity between average positive prediction rates among the two classes. The test will be a right-tailed hypothesis, since it is only a fairness violation if the positive prediction rate among white individuals is higher than the rate among black individuals. The results from the test are shown in the table below. 

\begin{table}[h!]

\caption{Figure 1.2}
\scalebox{1}{\begin{tabular}{|l|l|l|l|}
\hline
Model                        &  Disparity & Test Statistic & P-Value \\ \hline
Logistic Regression          & 0.023 & 31.76 & $~0$    \\ \hline
Linear Discriminant Analysis & 0.049 & 87.41 & $~0$   \\\hline
Random Forest           &      0.064 & 112.19 & $~0$    \\ \hline
Gradient Boosting            &  0.047 & 83.90 & $~0$  

\\ \hline
\end{tabular}}

\end{table}

All of the machine learning algorithms are therefore biased against black individuals based on the group fairness metric. In other words, there is a statistically significant (p $<$ 0.05) violation of the group fairness metric.

\subsection{Testing Predictive Parity}

Predictive Parity is a metric that tests whether individuals in both groups have an equal chance of truly belonging to the positive class given the fact that said individuals are predicted to fall into the negative class. Mathematically, this metrics tests whether $P(y=0|\hat{y}=0,race=white) = P(y=0|\hat{y}=0,race=black) $. K-Fold cross validation helps us obtain distributions for the metrics $P(y=0|\hat{y}=0,race=white)$ and $P(y=0|\hat{y}=0,race=black)$. A two sample t-test can then be used to detect whether there is a statistically significant disparity in positive predictive value (probability of being in the positive group given positive classification). The test will be a left tailed test since there is only a fairness violation if the average value of $P(y=0|\hat{y}=0,race=white)$ is less than the average value of $P(y=0|\hat{y}=0,race=black)$ as this would indicate that white individuals treated positively by the model would be more likely to act negatively compared to their black counterparts. The results from the t-tests are displayed in the table below.

\begin{table}[h!]

\caption{Figure 1.3}
\scalebox{1}{\begin{tabular}{|l|l|l|l|}
\hline
Model                        &  Disparity & Test Statistic & P-Value \\ \hline
Logistic Regression          & -0.050 & 44.06 & $~1$    \\ \hline
Linear Discriminant Analysis & -0.028 & 31.86 & $~1$   \\\hline
Random Forest           &      -0.012 & 14.33 & $~1$    \\ \hline
Gradient Boosting            &  -0.012 & 14.12 & $~1$  

\\ \hline
\end{tabular}}

\end{table}

Using all algorithms, a white individual assigned to the positive class has a higher probability of truly belonging to the positive class compared to a black individual assigned to the positive class. Therefore, there is no statistically significant violation of predictive parity in a manner that is unfair to black individuals. There is; however, reverse discrimination, as there is a statistically significant violation of predictive parity in a manner that is unfair to white individuals.

\subsection{Testing Predictive Equality}

Predictive equality is a metric that tests whether both the protected and unprotected groups have equivalent false positive rates. Mathematically, this metric tests whether $P(\hat{y}=0|y=1,race=white) =P(\hat{y}=0|y=1,race=black) $. K-Fold cross validation can be used to generate distributions for $P(\hat{y}=0|y=1,race=white)$ and $P(\hat{y}=0|y=1,race=black)$ and the means of these distributions can be compared using a two sample t-test. The tail will be a right tailed test as it is a fairness violation if white individuals are classified into the positive class while behaving negatively at a higher rate than black individuals. The results from the tests are shown in the table below.

\begin{table}[h!]

\caption{Figure 1.4}
\scalebox{1}{\begin{tabular}{|l|l|l|l|}
\hline
Model                        &  Disparity & Test Statistic & P-Value \\ \hline
Logistic Regression          & -0.001 & -2.03 & 0.979    \\ \hline
Linear Discriminant Analysis & 0.018 & 30.42 & $~0$  \\\hline
Random Forest           &      0.029 & 53.11 & $~0$    \\ \hline
Gradient Boosting            &  0.022 & 38.56 & $~0$  

\\ \hline
\end{tabular}}

\end{table}

All models, excluding logistic regression, exhibit significant bias against black individuals meaning that for these models, there is a statistically significant violation of predictive equality (p $<$ 0.05). For logistic regression, there seems to be reverse discrimination as black individuals have higher average false positive rates compared to white individuals and this disparity is statistically significant.

\subsection{Testing Equal Opportunity}

Equal Opportunity is a metric that tests whether the protected and unprotected groups have equivalent false negative rates. Mathematically, this metric tests whether $P(\hat{y}=1|y=0, race=white) = P(\hat{y}=1|y=0,race=black)$. K-Fold cross validation can be used to obtain distributions for $P(\hat{y}=1|y=0, race=white)$ and $P(\hat{y}=1|y=0,race=black)$ and a two sample t-test can be used to compare the means of the distributions. This test will be a left tailed test as it would be discrimination if a black individual had a higher probability than a white individual of being predicted to return to prison given the fact that said individual truly will not return to prison. The results from the tests are shown in the table below.

\begin{table}[h!]

\caption{Figure 1.5}
\scalebox{1}{\begin{tabular}{|l|l|l|l|}
\hline
Model                        &  Disparity & Test Statistic & P-Value \\ \hline
Logistic Regression          & 0.042 & -33.45 & $~0$    \\ \hline
Linear Discriminant Analysis & 0.071 & -76.79 & $~0$  \\\hline
Random Forest           &      0.088 & -93.31 & $~0$    \\ \hline
Gradient Boosting            &  0.054 & -64.18 & $~0$  

\\ \hline
\end{tabular}}

\end{table}

Based on the results, it is evident that there is a statistically significant violation of equal opportunity (p < 0.05) as with all models, black individuals have a higher false negative rate compared to their white counterparts.

\subsection{Testing Equalized Odds}

Equalized Odds is a metric that tests whether the protected and unprotected have equivalent true positive rates and false positive rates. Mathematically, this metric tests whether $P(\hat{y} = 0|y=0, race=white) = P(\hat{y} = 0|y=0, race=black)$ and if  $P(\hat{y} = 0|y=1, race=white) = P(\hat{y} = 0|y=1, race=black)$. K-Fold cross validation can be used to generate distributions for $P(\hat{y} = 0|y=0, race=white)$,$P(\hat{y} = 0|y=0, race=black)$, $P(\hat{y} = 0|y=1, race=white)$, and $P(\hat{y} = 0|y=1, race=black)$. Simultaneous two sample t-tests can be used to compare the means of the two sets of distributions. This test will use a significance level of 0.025 since two tests are being conducted simultaneously and therefore, the Bonferroni correction indicates that the significance level for each tests should be $\frac{0.05}{2}$ or 0.025. The comparison of true positive rates will be a right tailed test as it is discrimination if a black individual is less likely to be classified into the positive group while behaving positively compared to a white individual. The comparison of false positive rates will also be a right tailed test as it is discrimination if a white individual is more likely to be classified into the positive group despite behaving negatively compared to a black individual. The results from the tests are shown below.

\begin{table}[h!]

\caption{Figure 1.6 (FPR)}
\scalebox{1}{\begin{tabular}{|l|l|l|l|l|l|}
\hline
Model                        &  Disparity & Test Statistic & P-Value \\ \hline
Logistic Regression          & -0.001 & -2.03  & 0.979     \\ \hline
Linear Discriminant Analysis & 0.018& 30.42 & $~0$ \\\hline
Random Forest           &    0.029  & 53.11  & $~0$    \\ \hline
Gradient Boosting            &  0.022  & 38.56  & $~0$ 

\\ \hline
\end{tabular}}

\end{table}

\begin{table}[h!]

\caption{Figure 1.7 (TPR)}
\scalebox{1}{\begin{tabular}{|l|l|l|l|l|l|}
\hline
Model                        &  Disparity  & Test Statistic & P-Value  \\ \hline
Logistic Regression          &  0.042 & 33.45  & $~0$    \\ \hline
Linear Discriminant Analysis & 0.071  & 76.79 & $~0$  \\\hline
Random Forest           &   0.088 &  93.31 & $~0$   \\ \hline
Gradient Boosting            &  0.054  & 64.18  & $~0$ 

\\ \hline
\end{tabular}}

\end{table}

The tables show that under all models, at least one of the hypotheses can be rejected. Therefore, under all models, there is a statistically significant violation of the equalized odds metric.

\subsection{Testing Conditional Use Accuracy Equality}

Conditional use accuracy equality is a metric that tests whether the protected and unprotected groups have equal positive predictive values and negative predictive values. Mathematically, this metric tests whether $P(y=1|\hat{y}=1,race=white) = P(y=1|\hat{y}=1, race=black)$ and $P(y=0|\hat{y}=0,race=white) = P(y=0|\hat{y}=0,race=black)$. K-Fold cross validation can be used to generate the distributions for $P(y=1|\hat{y}=1,race=white)$,$P(y=1|\hat{y}=1, race=black)$,$P(y=0|\hat{y}=0,race=white)$, and $P(y=0|\hat{y}=0,race=black)$. Simultaneous two sample t-tests can then be used to compare the means of each set of distributions. The bonferroni correction states that the significance level for each test should be 0.025, and thus, that is the significance level that will be used for this test. The test for positive predictive values will be a left tailed test as it is discrimination if black individuals who are predicted to fall into the positive category have a higher chance of actually falling into the positive category compared to white individuals. The test for negative predictive values will be a right tailed test as it is discrimination if white individuals who are predicted to fall into the negative category actually fall into the negative category at a higher rate compared to white individuals. The results from the tests are shown below.

\begin{table}[h!]

\caption{Figure 1.8 (PPV)}
\scalebox{1}{\begin{tabular}{|l|l|l|l|l|l|}
\hline
Model                        &  Disparity & Test Statistic & P-Value \\ \hline
Logistic Regression          & -0.050 & 44.06 & $~1$    \\ \hline
Linear Discriminant Analysis & -0.028 & 31.86 & $~1$   \\\hline
Random Forest           &      -0.012 & 14.33 & $~1$    \\ \hline
Gradient Boosting            &  -0.012 & 14.12 & $~1$  

\\ \hline
\end{tabular}}

\end{table}

\begin{table}[h!]

\caption{Figure 1.9 (NPV)}
\scalebox{1}{\begin{tabular}{|l|l|l|l|l|l|}
\hline
Model                        &  Disparity  & Test Statistic & P-Value  \\ \hline
Logistic Regression          &  -0.004 & -5.20  & $~1$    \\ \hline
Linear Discriminant Analysis & 0.013  & 19.48 & $~0$  \\\hline
Random Forest           &   0.018 &  28.07 & $~0$   \\ \hline
Gradient Boosting            &  0.008  & 12.94  & $~0$ 

\\ \hline
\end{tabular}}

\end{table}

All models, excluding logistic regression, exhibit statistically significant disparities in negative predictive value among the groups. Therefore, all models, except logistic regression, violate the conditional use accuracy equality metric in a statistically significant manner.

\subsection{Testing Overall Accuracy Equality}

Overall accuracy equality is a metric that tests whether the protected and unprotected groups have identical prediction accuracy. Mathematically, this metric tests whether $P(y=\hat{y}, race=white) = P(y=\hat{y}|race=black)$. K-Fold cross validation can be used to generate distributions for $P(y=\hat{y}, race=white)$ and $P(y=\hat{y}|race=black)$ and a two sample t-test can be used to compare the means of the distributions. This will be a two sided test as a disparity in either direction would indicate discrimination. The results from this test are shown in the table below.

\begin{table}[h!]

\caption{Figure 1.10}
\scalebox{1}{\begin{tabular}{|l|l|l|l|l|l|}
\hline
Model                        &  Disparity  & Test Statistic & P-Value  \\ \hline
Logistic Regression          &  0.009 & 15.09  & $~0$    \\ \hline
Linear Discriminant Analysis & 0.016  & 29.92 & $~0$  \\\hline
Random Forest           &   0.015 &  28.86 & $~0$   \\ \hline
Gradient Boosting            &  0.007  & 14.24  & $~0$ 

\\ \hline
\end{tabular}}

\end{table}

The prediction accuracy for white individuals is higher than the prediction accuracy for black individuals under every model. This disparity is statistically significant for each model. Therefore, all models exhibit a statistically significant violation of overall accuracy equality.

\subsection{Testing Treatment Equality}

Treatment equality is a metric that tests whether the ratio of false negatives to false positives is the same in both the protected and unprotected groups. K-Fold cross validation can be used to generate a distribution of the ratio of false positives to false negatives for each group and a two sample t-test can be used to compare the means of the distribution. This will be a left tailed test as if the ratio is higher for black individuals compared to white individuals that will mean that black individuals are more often classified into the negative class when they truly belong to the positive class compared to when they are classified into the positive class while belonging to the negative class. The results from the tests are shown below.

\begin{table}[h!]

\caption{Figure 1.11}
\scalebox{1}{\begin{tabular}{|l|l|l|l|l|l|}
\hline
Model                        &  Disparity  & Test Statistic & P-Value  \\ \hline
Logistic Regression          &  -0.139 & 5.52 & 1    \\ \hline
Linear Discriminant Analysis & 0.296  & -21.87 & 0  \\\hline
Random Forest           &   0.579 &  -40.83 & 0   \\ \hline
Gradient Boosting            &  0.187 & -17.71 & 0

\\ \hline
\end{tabular}}

\end{table}

All models, aside from logistic regression, exhibit a significant disparity in ratio of false negatives to false positive. Furthermore, all models, except for logistic regression, significantly violate the treatment equality metric.

\subsection{Testing Calibration}

Calibration is a metric that tests whether for a given probability score the probability of an individual truly belonging to the positive class is equal for both the protected and unprotected groups. Mathematically, this metric tests whether $P(y=1|S=s,race=white) = P(y=1|S=s,race=black)$. K-Fold cross validation can be used to obtain predicted probabilities of recidivism for each observation in the dataset. We can then divide the probability scores into ten bins. We can then obtain the observed frequency of positive outcomes (no recidivism) within each bin for the unprotected class. We can then obtain the frequency of positive outcomes within each bin for the protected class and apply a standardization method to ensure that the frequencies of positive outcomes among both groups can be compared. The standardization method takes the percentage of positive outcomes in the bin for the protected group and multiplies that percentage by the total number of unprotected observations that fall within the bin. Let $\alpha$ be number of protected individuals in bin, let $\theta$ be number of protected individuals with positive outcomes in the bin, let $\beta$ be number of unprotected individuals in bin. Let $\lambda$ be the standardized frequency.

\begin{equation}
\lambda = \frac{\theta}{\alpha}\times\beta
\end{equation}

Now $\lambda$ can be compared with $\gamma$ which is the frequency of positive outcomes for the unprotected individuals using the chi-squared goodness of fit test. The null hypothesis states that $\lambda$ and $\gamma$ do not vary significantly across bins. The test statistic can be computed using the equation.

\begin{equation}
    \chi^2 = \sum_{i=1}^{k}\frac{(\lambda_i - \gamma_i)^2}{\gamma_i}
\end{equation}

K represents the total number of bins. The test statistic follows an $\chi^2$ distribution with k-1 degrees of freedom. The test will be a right tailed test as a high test statistic indicates deviation between the frequencies for the protected and unprotected groups. The results from the tests are shown in the table below.

\begin{table}[h!]

\caption{Figure 1.12}
\scalebox{1}{\begin{tabular}{|l|l|l|l|l|l|}
\hline
Model                        &  Chi Squared Test Statistic & P-Value  \\ \hline
Logistic Regression          &  47.70 & 0    \\ \hline
Linear Discriminant Analysis & 61.68 & 0  \\\hline
Random Forest           &   44.56 & 0   \\ \hline
Gradient Boosting            & 15.06 & 0.089

\\ \hline
\end{tabular}}

\end{table}

The probability predictions for all models, except for gradient boosting, violate the calibration definition in a statistically significant manner. Therefore, based on the calibration definition, the logistic regression, linear discriminant analysis, and random forest model are biased. 

\subsection{Testing Well Calibration}

Well calibration is a metric that extends the calibration metric to include criteria that states that for any predicted probability score s, the unprotected and protected groups must not only have equal probability of truly belonging to the positive class but that this probability should also be equal to the predicted probability score s. The test for calibration already checks whether $P(y=1|S=s,race=white) = P(y=1|S=s,race=black)$. Thus, to test for well calibration, the only criteria that remains to be tested is whether $P(y=1|S=s,race=white) \in s$  and $P(y=1|S=s,race=black) \in s$ where s is a bin of probability scores. A chi-square goodness of fit test can also be used to test this hypothesis. The null hypothesis would be $P(y=1|S=s,race=white) = P(y=1|S=s,race=black) = s$. Since s is a bin, when computing the test statistic from equation 2, $\gamma_i$ will be considered to be either the minimum or maximum value in the bin depending on whichever value minimizes the value of $\lambda_i - \gamma_i$. As a result, this test can be used to determine whether the observed frequencies for both the protected and unprotected groups deviate significantly from the expected frequency for each bin. Note that $\gamma_i$ will be multiplied by the total number of unprotected individuals with predicted probabilities that fall within the bin. Also note that the standardization formula in equation 1 will be used to compute $\lambda_i$. Thus, all frequencies compared will be standardized. The test statistic will be computed in the same manner as the test statistic for the calibration test but k will be multiplied by 2 as there are two comparisons being conducted in each bin. The results from the tests are shown in the table below. 

\begin{table}[h!]

\caption{Figure 1.13}
\scalebox{1}{\begin{tabular}{|l|l|l|l|l|l|}
\hline
Model                        &  Chi Squared Test Statistic & P-Value  \\ \hline
Logistic Regression          &  59.56 & 0    \\ \hline
Linear Discriminant Analysis & 62.61 & 0  \\\hline
Random Forest           &    54.27 & 0   \\ \hline
Gradient Boosting            & 27.94 & 0.084

\\ \hline
\end{tabular}}

\end{table}

The probability predictions for all models, except for gradient boosting, violate the well calibration definition in a statistically significant manner. Therefore, based on the well calibration definition, the logistic regression, linear discriminant analysis, and random forest model are biased. 

\subsection{Testing Balance for Positive Class}

Balance for Positive Class is a metric that tests whether the expected  probability score among individuals who truly belong to the positive class is the same between the protected and unprotected groups. Mathematically, this metric tests whether $E(S|y=0,race=white) = E(S|y=0,race=black)$. K-Fold cross validation can be used to generate distributions for $E(S|y=0,race=white)$ and $=E(S|y=0,race=black)$ and then a two sample t-test can be used to compare means of the two distributions. The test will be a left tailed test as it is discrimination if a black individual that truly will not return to prison is assigned a higher average probability of returning to prison compared to a white individual that truly will not return to prison. The results from the test are shown below.

\begin{table}[h!]

\caption{Figure 1.14}
\scalebox{1}{\begin{tabular}{|l|l|l|l|l|l|}
\hline
Model                        &  Disparity  & Test Statistic & P-Value  \\ \hline
Logistic Regression          &  0.021 & -52.08 & 0    \\ \hline
Linear Discriminant Analysis & 0.042  & -101.02 & 0  \\\hline
Random Forest           &   0.043 &  -121.74 & 0   \\ \hline
Gradient Boosting            &  0.038 & -78.93 & 0

\\ \hline
\end{tabular}}

\end{table}

The results show that for each model, there is a statistically significant disparity in expected probability scores among individuals with positive outcomes in both groups.

\subsection{Testing Balance for Negative Class}

Balance for Negative Class is a metric that tests whether the expected  probability score among individuals who truly belong to the negative class is the same between the protected and unprotected groups. Mathematically, this metric tests whether $E(S|y=1,race=white) = E(S|y=1,race=black)$. K-Fold cross validation can be used to generate distributions for $E(S|y=1,race=white)$ and $=E(S|y=1,race=black)$ and then a two sample t-test can be used to compare means of the two distributions. The test will be a left tailed test as it is discrimination if a white individual that truly will return to prison is assigned a lower average probability of returning to prison compared to a black individual that truly will return to prison. The results from the test are shown below.

\begin{table}[h!]

\caption{Figure 1.15}
\scalebox{1}{\begin{tabular}{|l|l|l|l|l|l|}
\hline
Model                        &  Disparity  & Test Statistic & P-Value  \\ \hline
Logistic Regression          &  0.123 & -354.25 & 0    \\ \hline
Linear Discriminant Analysis & 0.259  & -686.14 & 0  \\\hline
Random Forest           &   0.218 &  -686.37 & 0   \\ \hline
Gradient Boosting            &  0.321 & -732.08 & 0

\\ \hline
\end{tabular}}

\end{table}

The results show that for each model, there is a statistically significant disparity in expected probability scores among individuals with negative outcomes in both groups.

\subsection{Testing Causal Discrimination}

Causal Discrimination tests whether individuals with identical attributes aside from race have identical predictions. Mathematically, causal discrimination would test whether $f((X, A = 1)) - f((X, A = 0)) = 0$ where $X$ represents covariates that are held constant, $A$ represents racial group, and $f$ represents predicted value based on input data vector. To test whether racial group has a statistically significant causal effect on predicted outcome, K-Fold cross validation can be used to generate predictions for each observation in the dataset and then a race change can be simulated to generate counterfactual predictions for each observation. The agreements between the true predictions and the simulated predictions for each model are depicted in the tables below.

\begin{table}[h!]

\caption{Figure 1.16 (Logistic Regression)}
\scalebox{1}{\begin{tabular}{|l|l|l|l|l|l|}
\hline
Original (X) vs. Simulated (Y)  for Unprotected Group                 &   0 & 1  \\ \hline
0        &  3029 & 13    \\ \hline
1 & 15  & 7892   \\\hline
Original (X) vs. Simulated (Y)  for Protected Group                 &   0 & 1  \\ \hline
0        &  3968 & 27    \\ \hline
1 & 14  & 10792  \\\hline

\end{tabular}}

\end{table}

\begin{table}[h!]

\caption{Figure 1.17 (Linear Discriminant Analysis)}
\scalebox{1}{\begin{tabular}{|l|l|l|l|l|l|}
\hline

Original (X) vs. Simulated (Y)  for Unprotected Group                 &   0 & 1  \\ \hline
0        &  4257 & 265   \\ \hline
1 & 0  & 6427   \\\hline
Original (X) vs. Simulated (Y)  for Protected Group                 &   0 & 1  \\ \hline
0        &  4673 & 0    \\ \hline
1 & 354  & 9774  \\\hline

\end{tabular}}

\end{table}

\begin{table}[h!]

\caption{Figure 1.18 (Random Forest Algorithm)}
\scalebox{1}{\begin{tabular}{|l|l|l|l|l|l|}
\hline
Original (X) vs. Simulated (Y)  for Unprotected Group                 &   0 & 1  \\ \hline
0        &  3922 & 136    \\ \hline
1 & 153  & 6738   \\\hline
Original (X) vs. Simulated (Y)  for Protected Group                 &   0 & 1  \\ \hline
0        &  4399 & 208    \\ \hline
1 & 184  & 10010  \\\hline

\end{tabular}}

\end{table}

\begin{table}[h!]

\caption{Figure 1.19 (Extreme Gradient Boosting)}
\scalebox{1}{\begin{tabular}{|l|l|l|l|l|l|}
\hline
Original (X) vs. Simulated (Y)  for Unprotected Group                 &   0 & 1  \\ \hline
0        &  4405 & 230    \\ \hline
1 & 48  & 6266   \\\hline
Original (X) vs. Simulated (Y)  for Protected Group                 &   0 & 1  \\ \hline
0        &  4971 & 84    \\ \hline
1 & 355  & 9391  \\\hline
\end{tabular}}

\end{table}

These contingency tables can be used to test for causal discrimination using McNemar's test. McNemar's test identifies 
whether there are equal proportions of both types of discordant pairs in the contingency table. In this case, the test will identify whether for unprotected or protected individuals, the proportion of observations where the original prediction was 1 and the counterfactual prediction was 0 is equivalent to the proportion of observations where the original prediction was 0 and the counterfactual prediction was 1. In some of the tables, we can see that there are very few "discordant" pairs and as a result, the test may not have enough power. Therefore, the Mcnemar mid-p test should be used to ensure adequate power. The test has a p-value that is 

\begin{equation}
    2 \sum_{i=0}^{k}\binom{n}{i}0.5^i (1-0.5)^{n-i}-\binom{n}{k}0.5^k(1-0.5)^{n-k}
\end{equation}

Note that n represents the total number of discordant pairs, and k represents the number of discordant pairs where the original prediction is 1 and the counterfactual prediction is 0. If the p-value is less than 0.05 the null hypothesis can be rejected and we can infer that the proportions of each type of discordant pairs are not equivalent.The results from the tests are shown below.

\begin{table}[h!]

\caption{Figure 1.20}
\scalebox{1.05}{\begin{tabular}{|l|l|l|l|l|l|}
\hline
Model                        &  K & N-K & P-Value  \\ \hline
Logistic Regression (Unprotected)       &  13 & 15 & 0.78    \\ \hline
Logistic Regression (Protected)       &  27 & 14 & 0.0516    \\ \hline

Linear Discriminant Analysis (Unprotected) & 265 & 0 & 0  \\\hline

Linear Discriminant Analysis (Protected) & 0 & 354 & 0  \\\hline
Random Forest (Unprotected)           &   136 &  153 & 0.332   \\ \hline
Random Forest (Protected)           &   208 &  184 & 0.236   \\ \hline
Gradient Boosting (Unprotected)         &  230 & 48 & 0

\\ \hline

Gradient Boosting (Unprotected)         &  84 & 355 & 0

\\ \hline
\end{tabular}}

\end{table}

Based on the table above, the conclusion can be made that the for the counterfactual predictions of the linear discriminant analysis and gradient boosting algorithms, the proportions of each type of discordant pairs are not equivalent. However, this is not an indication of bias as for both models, the counterfactual for a white individual is treated more positively compared to the white individual, and similarly, the counterfactual for a black individual is treated more negatively compared to the black individual. Therefore, these models are not discriminatory and instead may be practicing reverse discrimination.

\subsection{Testing Fairness Through Awareness}

Fairness through Awareness is a metric that tests the principle that two individuals with similar attributes should be classified into similar groups. The dataset of prisoners contains data points $(X,A,\hat{Y})$ where $X$ represents a set of covariates, $A$ represents the racial group, and $\hat{Y}$ represents the predicted outcome. K-Fold cross validation can be used to generate predictions for each individual in the protected and unprotected classes. After this, each protected individual can be matched with their closest privileged counterpart and the predicted recidivism outcomes for both individuals can be compared. The mahalanobis distance is used to match protected and privileged individuals.

The mahalanobis distance is computed using the following formula.

\begin{equation}
    M(x,y) = \sqrt{(x-y)^{T}S^{-1}(x-y)}
\end{equation}

$M(x,y)$ represents the mahalanobis distance between the vectors $x$ and $y$. Note that $S$ is the covariance matrix between the two vectors. After applying the algorithm above, we can compare the observations from both the unprotected and protected groups with their nearest neighbors using contingency tables. The contingency tables for each group under each model are shown below.

\begin{table}[h!]

\caption{Figure 1.21 (Logistic Regression)}
\scalebox{1}{\begin{tabular}{|l|l|l|l|l|l|}
\hline
Original (X) vs. Neighbor (Y)  for Unprotected Group                 &   0 & 1  \\ \hline
0        &  641 & 2167    \\ \hline
1 & 2403  & 5738   \\\hline
Original (X) vs. Neighbor (Y)  for Protected Group                 &   0 & 1  \\ \hline
0        &  807 & 2727    \\ \hline
1 & 3175  & 8022  \\\hline

\end{tabular}}

\end{table}

\begin{table}[h!]

\caption{Figure 1.22 (Linear Discriminant Analysis)}
\scalebox{1}{\begin{tabular}{|l|l|l|l|l|l|}
\hline

Original (X) vs. Neighbor (Y)  for Unprotected Group                 &   0 & 1  \\ \hline
0        &  1252 & 2337   \\ \hline
1 & 3005  & 4355   \\\hline
Original (X) vs. Neighbor (Y)  for Protected Group                 &   0 & 1  \\ \hline
0        &  1661 & 3723    \\ \hline
1 & 3366 & 6051 \\\hline

\end{tabular}}

\end{table}

\begin{table}[h!]

\caption{Figure 1.23 (Random Forest Algorithm)}
\scalebox{1}{\begin{tabular}{|l|l|l|l|l|l|}
\hline
Original (X) vs. Neighbor (Y)  for Unprotected Group                 &   0 & 1  \\ \hline
0        &  1070 & 2105    \\ \hline
1 & 3005  & 4769   \\\hline
Original (X) vs. Neighbor (Y)  for Protected Group                 &   0 & 1  \\ \hline
0        &  1380 & 3654   \\ \hline
1 & 3203  & 6564  \\\hline

\end{tabular}}

\end{table}

\begin{table}[h!]

\caption{Figure 1.24 (Extreme Gradient Boosting)}
\scalebox{1}{\begin{tabular}{|l|l|l|l|l|l|}
\hline
Original (X) vs. Neighbor (Y)  for Unprotected Group                 &   0 & 1  \\ \hline
0        &  1491 & 2357    \\ \hline
1 & 2962  & 6564  \\\hline
Original (X) vs. Neighbor (Y)  for Protected Group                 &   0 & 1  \\ \hline
0        &  1879 & 3726    \\ \hline
1 & 3447  & 5749  \\\hline
\end{tabular}}

\end{table}

The McNemar's mid-p test that was used for testing Causal Discrimination can be used to test for Fairness Through Awareness as these metrics test similar principles since causal discrimination tests whether identical individuals will be classified differently while fairness through awareness tests whether similar individuals will be classified differently. The results from the test are shown in the table below.

\begin{table}[h!]

\caption{Figure 1.20}
\scalebox{1.05}{\begin{tabular}{|l|l|l|l|l|l|}
\hline
Model                        &  K & N-K & P-Value  \\ \hline
Logistic Regression (Unprotected)       &  2167 & 2403 & 0.0005    \\ \hline
Logistic Regression (Protected)       &  2727 & 3175 & 0   \\ \hline

Linear Discriminant Analysis (Unprotected) & 2337 & 3005 & 0  \\\hline

Linear Discriminant Analysis (Protected) & 3723 & 3366 & 0  \\\hline
Random Forest (Unprotected)           &   2105 &  3005 & 0   \\ \hline
Random Forest (Protected)           &   3654 &  3203 & 0   \\ \hline
Gradient Boosting (Unprotected)         &  2357 & 2962 & 0

\\ \hline

Gradient Boosting (Unprotected)          &  3726 & 3447 & 0.001

\\ \hline
\end{tabular}}

\end{table}

Under each model, the test for each group returns a statistically significant result. For each model, the closest black individual to a white individual will be treated more negatively than the white individual and the closest white individual to a black individual will be treated more positively than the black individual. Therefore, for each of the four models, there is a statistically significant violation of the fairness through awareness principle. Furthermore, under this definition, all of the models are biased against black individuals.

\section*{Conclusion}

This paper established a framework that can be to test machine learning algorithms for fairness violations. Statistical tests were introduced to test an algorithms' compliance with many commonly used fairness definitions. This framework was then tested through an experiment in which four machine learning algorithms were trained to predict recidivism outcomes. The algorithms were then rigorously tested using the developed framework. Most models violated nearly all fairness metrics that rely on comparing predicted and actual outcomes as well as all metrics that rely on comparing predicted probabilities and actual outcomes. When it came to metrics based on the principle that similar observations should be classified similarly; however, some models showcased reverse discrimination when individuals with identical attributes but different races were compared. All models did exhibit discrimination in the sense that black individuals were treated more negatively that their white counterpart with the most similar attributes. Therefore, under many fairness definitions, the models trained to predict recidivism outcomes could be considered discriminatory against black individuals. However, under other fairness definitions, the models trained to predict recidivism could considered discriminatory against white individuals. Therefore, future research in this field should discuss establishing more strict definitions by which models can be tested for algorithmic bias to avoid the issue of different definitions leading to different conclusions.

\vspace{12pt}

\end{document}